\documentclass[runningheads]{llncs}
\usepackage{graphicx}
\usepackage[english]{babel}
\usepackage[utf8]{inputenc}
\usepackage[caption=false,font=footnotesize]{subfig}
\usepackage[export]{adjustbox}
\usepackage{caption}
\usepackage{booktabs}
\usepackage{amsmath,amssymb}
\usepackage[pagebackref=true]{hyperref} 

\usepackage{collcell}
\usepackage{hhline}
\usepackage{pgf}
\usepackage{multirow}
\usepackage{blindtext}
\usepackage{enumitem}
\usepackage{bbm}
\usepackage{siunitx}
\usepackage{arydshln}

\usepackage{accents}

\usepackage{wrapfig}
\usepackage{bm}
\usepackage{soul}
\usepackage[normalem]{ulem}
\usepackage{makecell}
\def\colorModel{hsb} %
\newcommand\ColCell[1]{
  \pgfmathparse{#1<50?1:0}  %
    \ifnum\pgfmathresult=0\relax\color{white}\fi
  \pgfmathsetmacro\compA{0}      %
  \pgfmathsetmacro\compB{#1/100} %
  \pgfmathsetmacro\compC{1}      %
  \edef\x{\noexpand\centering\noexpand\cellcolor[\colorModel]{\compA,\compB,\compC}}\x #1
  } 
\newcolumntype{E}{>{\collectcell\ColCell}m{0.45cm}<{\endcollectcell}}  %

\usepackage[T1]{fontenc} 
\usepackage{lipsum}

\usepackage{tcolorbox}
\newtcolorbox{afancybox}[1][]{#1,colback=white,colframe=black}
\usepackage{ulem}

\usepackage{commath}
\usepackage[misc,geometry]{ifsym}
\usepackage[title]{appendix}
\usepackage[figuresright]{rotating}
\usepackage{booktabs}
\usepackage{nicefrac}

\newcommand{\z}{\mathbf{z}}
\newcommand{\p}{\mathbf{p}}
\newcommand{\LL}{\mathcal{L}}

\newcommand{\unl}[1]{\underline{#1}}
\newcommand{\ft}[1]{\underline{\textbf{#1}}}
\newcommand{\sd}[1]{\textbf{#1}}
\newcommand{\xdownarrow}[1]{%
  {\left\downarrow\vbox to #1{}\right.\kern-\nulldelimiterspace}
}

\begin{document}

\title{Multi-Head Multi-Loss Model Calibration}

\titlerunning{Multi-Head Multi-Loss Model Calibration}
\author{Adrian Galdran\inst{1,2}$^\textrm{,\Letter}$ \and
Johan Verjans\inst{2} \and \\
Gustavo Carneiro\inst{2} \and
Miguel A. González Ballester\inst{1,3}}
\authorrunning{A. Galdran et al.}
\institute{BCN Medtech, Universitat Pompeu Fabra, Barcelona, Spain, \email{\{adrian.galdran,ma.gonzalez\}@upf.edu}
\and
AIML, University of Adelaide, Australia, \email{johan.verjans@adelaide.edu}
\and
University of Surrey, Guildford, UK, \email{g.carneiro@surrey.ac.uk}
\and
Catalan Institution for Research and Advanced Studies (ICREA), Barcelona, Spain }

\maketitle              %
\begin{abstract}
Delivering meaningful uncertainty estimates is essential for a successful deployment of machine learning models in the clinical practice. 
A central aspect of uncertainty quantification is the ability of a model to return predictions that are well-aligned with the actual probability of the model being correct, also known as model calibration. 
Although many methods have been proposed to improve calibration, no technique can match the simple, but expensive approach of training an ensemble of deep neural networks. 
In this paper we introduce a form of simplified ensembling that bypasses the costly training and inference of deep ensembles, yet it keeps its calibration capabilities. 
The idea is to replace the common linear classifier at the end of a network by a set of heads that are supervised with different loss functions to enforce diversity on their predictions. 
Specifically, each head is trained to minimize a weighted Cross-Entropy loss, but the weights are different among the different branches. 
We show that the resulting averaged predictions can achieve excellent calibration without sacrificing accuracy in two challenging datasets for histopathological and endoscopic image classification.
Our experiments indicate that Multi-Head Multi-Loss classifiers are inherently well-calibrated, outperforming other recent calibration techniques and even challenging Deep Ensembles' performance.
Code to reproduce our experiments can be found at \url{https://github.com/agaldran/mhml_calibration} .
\end{abstract}

\setcounter{footnote}{0}

\section{Introduction and Related Work}
When training supervised computer vision models, we typically focus on improving their predictive performance, yet equally important for safety-critical tasks is their ability to express meaningful uncertainties about their own predictions \cite{chua_tackling_2022}. 
In the context of machine learning, we often distinguish two types of uncertainties: \textit{epistemic} and \textit{aleatoric} \cite{hullermeier_aleatoric_2021}. 
Briefly speaking, epistemic uncertainty arises from imperfect knowledge of the model about the problem it is trained to solve, whereas aleatoric uncertainty describes ignorance regarding the data used for learning and making predictions. 
For example, if a classifier has learned to predict the presence of cancerous tissue on a colon histopathology, and it is tasked with making a prediction on a breast biopsy it may display epistemic uncertainty, as it was never trained for this problem \cite{linmans_predictive_2023}. 
Nonetheless, if we ask the model about a colon biopsy with ambiguous visual content, \textit{i.e.} a hard-to-diagnose image, then it could express aleatoric uncertainty, as it may not know how to solve the problem, but the ambiguity comes from the data. 
This distinction between epistemic and aleatoric is often blurry, because the presence of one of them does not imply the absence of the other \cite{hullermeier_quantifying_2022}. 
Also, under strong epistemic uncertainty, aleatoric uncertainty estimates can become unreliable \cite{valdenegro-toro_deeper_2022}.

Producing good uncertainty estimates can be useful, \textit{e.g.} to identify test samples where the model predicts with little confidence and which should be reviewed \cite{bernhardt_failure_2022}. 
A straightforward way to report uncertainty estimates is by interpreting the output of a model (maximum of its softmax probabilities) as its predictive confidence. 
When this confidence aligns with the actual accuracy we say that the model is calibrated \cite{filho_classifier_2021}. 
Model calibration has been studied for a long time, with roots going back to the weather forecasting field \cite{brier_verification_1950}. 
Initially applied mostly for binary classification systems \cite{ferrer_analysis_2022}, the realization that modern neural networks tend to predict over-confidently \cite{guo_calibration_2017} has led to a surge of interest in recent years \cite{filho_classifier_2021}. 
Broadly speaking, one can attempt to promote calibration during training, by means of a post-processing stage, or by model ensembling.

\paragraph{\textbf{Training-Time Calibration}}
Popular training-time approaches consist of reducing the predictive entropy by means of regularization \cite{hebbalaguppe_stitch_2022}, \textit{e.g.} Label Smoothing \cite{muller_when_2019} or MixUp \cite{thulasidasan_mixup_2019}, or loss functions that smooth predictions \cite{mukhoti_calibrating_2020}. 
These techniques often rely on correctly tuning a hyper-parameter controlling the trade-off between discrimination ability and confidence, and can easily achieve better calibration at the expense of decreasing predictive performance \cite{liu_devil_2022}. 
Examples of medical image analysis works adopting this approach are Difference between Confidence and Accuracy regularization \cite{liang_improved_2020} for medical image diagnosis, or Spatially-Varying and Margin-Based Label Smoothing \cite{islam_spatially_2021,murugesan_calibrating_2022}, which extend and improve Label Smoothing for biomedical image segmentation tasks.

\paragraph{\textbf{Post-Hoc Calibration}}
Post-hoc calibration techniques like Temperature Scaling \cite{guo_calibration_2017} and its variants \cite{ding_local_2021,kull_beyond_2019} have been proposed to correct over or under-confident predictions by applying simple monotone mappings (fitted on a held-out subset of the training data) on the output probabilities of the model. 
Their greatest shortcoming is the dependence on the \textit{i.i.d.} assumption implicitly made when using validation data to learn the mapping: these approaches suffer to generalize to unseen data \cite{ovadia_can_2019}. 
Other than that, these techniques can be combined with training-time methods and return compounded performance improvements.

\paragraph{\textbf{Model Ensembling}}
A third approach to improve calibration is to aggregate the output of several models, which are trained beforehand so that they have some diversity in their predictions \cite{dietterich_ensemble_2000}. 
In deep learning, model ensembles are considered to be the most successful method to generate meaningful uncertainty estimates \cite{lakshminarayanan_simple_2017}. 
An obvious weakness of deep ensembles is the requirement of training and then keeping for inference purposes a set of models, which results in a computational overhead that can be considerable for larger architectures. 
Examples of applying ensembling in medical image computing include \cite{larrazabal_orthogonal_2021,ma_test-time_2022}.

In this work we achieve model calibration by means of multi-head models trained with diverse loss functions. 
In this sense, our approach is closest to some recent works on multi-output architectures like \cite{linmans_predictive_2023}, where a multi-branch CNN is trained on histopathological data, enforcing specialization of the different heads by backpropagating gradients through branches with the lowest loss. 
Compared to our approach, ensuring correct gradient flow to avoid dead heads requires ad-hoc computational tricks \cite{linmans_predictive_2023}; in addition, no analysis on model calibration on in-domain data or aleatoric uncertainty was developed, focusing instead on anomaly detection. 
Our main \textbf{contribution} is a multi-head model that \textbf{I)} exploits multi-loss diversity to achieve greater confidence calibration than other learning-based methods, while \textbf{II)} avoiding the use of training data to learn post-processing mappings as most post-hoc calibration methods do, 
and \textbf{III)} sidesteping the computation overhead of deep ensembles. 

\section{Calibrated Multi-Head Models}
In this section we formally introduce multi-head models \cite{lee_why_2015}, and justify the need for enforcing diversity on them. 
Detailed derivations of all the results below are provided in the online supplementary materials.

\subsection{Multi-Head Ensemble Diversity}\label{mhd}
Consider a $K$-class classification problem, and a neural network $U_\theta$ taking an image $\mathbf{x}$ and mapping it onto a representation $U_\theta(\mathbf{x})\in\mathbb{R^N}$, which is linearly transformed by $f$ into a logits vector $\z=f(U_\theta(\mathbf{x}))\in\mathbb{R^K}$. 
This is then mapped into a vector of probabilities $\p\in [0,1]^K$ by a softmax operation $\p=\sigma(\z)$, where $p_j=e^{z_j}/\sum_i e^{z_i}$. 
If the label of $\mathbf{x}$ was $y\in\{1,...,K\}$, we can measure the error associated to prediction $\p$ with the cross-entropy loss $\LL_{\textrm{CE}}(\p, y)=-\log(p_y)$. 

\begin{figure}[htp!]
\centerline{\includegraphics[width=1\textwidth]{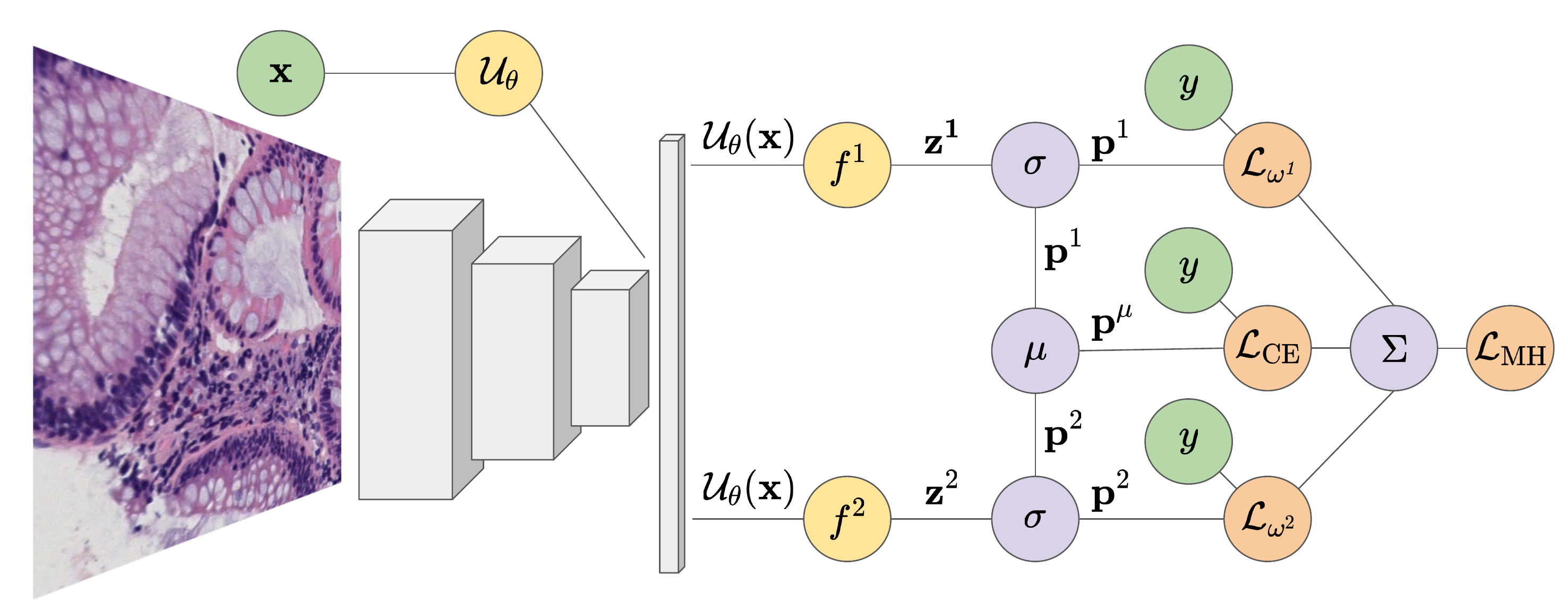}}
\caption{A multi-head multi-loss model with $M$=2 heads. 
An image $\mathbf{x}$ goes through a neural network $\mathcal{U}_\theta$ and then is linearly transformed by $M$ heads $\{f^m\}_{m=1}^M$, followed by softmax operations $\sigma$, into probability vectors $\{\p^m\}_{m=1}^M$. The final loss $\LL_{\textrm{MH}}$ is the sum of per-head weighted-CE losses $\mathcal{L}_{\boldsymbol{\omega}^m\textrm{-CE}}(\p^m,y)$ and the CE loss $\mathcal{L}_{\textrm{CE}}(\p^\mu,y)$ of the average prediction $\p^\mu=\mu(\p^1,...,\p^m)$. We modify the weights $\boldsymbol{\omega}^m$ between branches to achieve more diverse gradients during training.}
\label{fig_overview}
\end{figure}

We now wish to implement a multi-head ensemble model like the one shown in Fig. \ref{fig_overview}. For this, we replace $f$ by $M$ different branches $f^1,..., f^M$, each of them still taking the same input but mapping it to different logits $\z^m = f^m(U_\theta(\mathbf{x}))$. 
The resulting probability vectors $\p^m=\sigma(\z^m)$ are then averaged to obtain a final prediction $\p^\mu = (1/M)\sum_m \p^m$. 
We are interested in backpropagating the loss $\LL_{\textrm{CE}}(\p^\mu, y)=-\log(p^\mu_y)$ to find the gradient at each branch, $\nabla_{\z^m}\LL_{\textrm{CE}}(\p^\mu, y)$.

\paragraph{\textbf{Property 1:}}
For the M-head classifier in Fig. \ref{fig_overview}, the derivative of the cross-entropy loss at head $f^m$ with respect to $\z^m$ is given by 
\begin{equation}\label{multi_head_grad}
\nabla_{\z^m} \LL_{\textrm{CE}}(\p^\mu, y) = \frac{p_y^m}{\sum_i p_y^i}(\p^\mu - \mathbf{y}),
\end{equation}
where $\mathbf{y}$ is a one-hot representation of the label $y$.

From eq. (\ref{multi_head_grad}) we see that the gradient in branch $m$ will be scaled depending on how much probability mass $p_y^m$ is placed by $f^m$ on the correct class relative to the total mass placed by all heads. 
In other words, if every head learned to produce a similar prediction (not necessarily correct) for a particular sample, then the optimization process of this network would result in the same updates for all of them. 
As a consequence, diversity in the predictions that make up the output $\p^\mu$ of the network would be damaged.

\subsection{Multi-Head Multi Loss Models}\label{mhml}
In view of the above, one way to obtain more diverse gradient updates in a multi-head model during training could be to supervise each head with a different loss function. 
To this end, we will apply the weighted cross-entropy loss, given by $\mathcal{L}_{\boldsymbol{\omega}\textrm{-CE}}(\p, y) = -\omega_y \log(p^\mu_y)$, where $\boldsymbol{\omega}\in\mathbb{R}^K$ is a weight vector. 
In our case, we assign to each head a different weight vector $\boldsymbol{\omega}^m$ (as detailed below), in such a way that a different loss function $\mathcal{L}_{\boldsymbol{\omega^m}\textrm{-CE}}$ will supervise the intermediate output of each branch $f^m$, similar to deep supervision strategies \cite{lee_deeply-supervised_2015} but enforcing diversity. 
The total loss of the complete model is the addition of the per-head losses and the overall loss acting on the average prediction:
\begin{equation}\label{overall}
\LL_{\textrm{MH}}(\p, y) = \LL_{\textrm{CE}}(\p^\mu, y) + \sum_{m=1}^M \mathcal{L}_{\boldsymbol{\omega}^m\textrm{-CE}}(\p^m, y),
\end{equation}
where $\p=(\p^1,...,\p^M)$ is an array collecting all the predictions the network makes.
Since $\mathcal{L}_{\boldsymbol{\omega}\textrm{-CE}}$ results from just multiplying by a constant factor the conventional CE loss, we can readily calculate the gradient of $\LL_{\textrm{MH}}$  at each branch.

\paragraph{\textbf{Property 2:}}
For the Multi-Loss Multi-Head classifier shown in Fig. \ref{fig_overview}, the gradient of the Multi-Head loss $\LL_{\textrm{MH}}$ at branch $f^m$ is given by: 
\begin{equation}\label{overall_grad}
\nabla_{\z^m} \LL_{\textrm{MH}}(\p, y) = \left(\omega_y^m + \frac{p_y^m}{\sum_i p_y^i}\right)(\p^\mu - \mathbf{y}).
\end{equation}
Note that having equal weight vectors in all branches fails to break the symmetry in the scenario of all heads making similar predictions. 
Indeed, if for any two given heads $f^{m_i}, f^{m_j}$ we have $\boldsymbol{\omega}^{m_i}=\boldsymbol{\omega}^{m_j}$ and $\p^{m_i} \approx \p^{m_j}$, \textit{i.e.} $\p^m \approx \p^\mu \ \forall m$, then the difference in norm of the gradients of two heads would be:
\begin{equation}
\displaystyle\|\nabla_{\z^{m_i}} \LL_{\textrm{MH}}(\p, y) - \nabla_{\z^{m_j}} \LL_{\textrm{MH}}(\p, y)\|_1 \approx |\omega_y^{m_i} - \omega_y^{m_j}| \cdot \|\p^\mu-\mathbf{y}\|_1 = 0.
\end{equation}
It follows that we indeed require a different weight in each branch. 
In this work, we design a weighting scheme to enforce the specialization of each head into a particular subset of the categories $\{c_1, ..., c_K\}$ in the training set. 

We first assume that the multi-head model has less branches than the number of classes in our problem, \textit{i.e.} $M\leq K$, as otherwise we would need to have different branches specializing in the same category. 
In order to construct the weight vector $\boldsymbol{\omega}^m$, we associate to branch $f^m$ a subset of $N/K$ categories, randomly selected, for specialization, and these are weighed with $\omega^m_j=K$. 
Then, the remaining categories in $\boldsymbol{\omega}^m$ receive a weight of $\omega^m_j=1/K$. 
For example, in a problem with $4$ categories and $2$ branches, we could have $\boldsymbol{\omega}^1=[2,\nicefrac{1}{2},2,\nicefrac{1}{2}]$ and $\boldsymbol{\omega}^2=[\nicefrac{1}{2},2,\nicefrac{1}{2},2]$.
If $N$ is not divisible by $K$, the reminder categories are assigned for specialization to random branches.

\subsection{Model Evaluation}
When measuring model calibration, the standard approach relies on observing the test set accuracy at different confidence bands $B$. 
For example, taking all test samples that are predicted with a confidence around $c=0.8$, a well-calibrated classifier would show an accuracy of approximately $80\%$ in this test subset. 
This can be quantified by the Expected Calibration Error (ECE), given by:
\begin{equation}
\mathrm{ECE} = \sum_{s=1}^N \frac{|B_s|}{N} |\mathrm{acc}(B_s) - \mathrm{conf}(B_s)|,
\end{equation}
where $ \bigcup_s B_s$ form a uniform partition of the unit interval, and $\mathrm{acc}(B_s)$, $\mathrm{conf}(B_s)$ are accuracy and average confidence (maximum softmax value) for test samples predicted with confidence in $B_s$.

In practice, the ECE alone is not a good measure in terms of practical usability, as one can have a perfectly ECE-calibrated model with no predictive power \cite{reinke_understanding_2023}.
A binary classifier in a balanced dataset, randomly predicting always one class with $c=0.5+\epsilon$ confidence, has a perfect calibration and $50\%$ accuracy. 
Proper Scoring Rules like Negative Log-Likelihood (NLL) or the Brier score are alternative uncertainty quality metrics \cite{gneiting_strictly_2007} that capture both discrimination ability and calibration: a model must be \textit{both accurate and calibrated} to achieve a low PSR value.
We report NLL, and also standard Accuracy, which contrary to ECE can be high even for badly-calibrated models. 
Finally, we show as summary metric the average rank when aggregating rankings of ECE, NLL, and accuracy.

\section{Experimental Results}
We now describe the data we used for experimentation, carefully analyze performance for each dataset, and end up with a discussion of our findings.

\subsection{Datasets and Architectures}
We conducted experiments on two datasets: \textbf{1)} the \textbf{Chaoyang} dataset\footnote{https://bupt-ai-cz.github.io/HSA-NRL/}, which contains colon histopathology images. 
It has 6,160 images unevenly distributed in 4 classes (29\%, 19\%, 37\%, 15\%), with some amount of label ambiguity, reflecting high aleatoric uncertainty. 
As a consequence, the best model in the original reference \cite{zhu_hard_2022}, applying specific techniques to deal with label noise, achieved an accuracy of 83.4\%. \textbf{2)} \textbf{Kvasir}\footnote{https://datasets.simula.no/hyper-kvasir/},  a dataset for the task of endoscopic image classification. The annotated part of this dataset contains 10,662 images, and it represents a challenging classification problem due a high amount of classes (23) and highly imbalanced class frequencies \cite{borgli_hyperkvasir_2020}.
For the sake of readability we do not show measures of dispersion, but we add them to the supplementary material (Appendix \ref{more_results}), together with further experiments on other datasets.

We implement the proposed approach by optimizing several popular neural network architectures, namely a common ResNet50 and two more recent models: a ConvNeXt \cite{liu_swin_2021} and a Swin-Transformer \cite{liu_swin_2021}. 
All models are trained for 50 epochs, which was observed enough for convergence, using Stochastic Gradient Descent with a learning rate of $l=1e$-$2$. 
Code to reproduce our results and hyperparameter specifications are shared at \url{github.com/withheld}.

\subsection{Performance Analysis}\label{perf_sec}
\subsubsection*{Notation:} 
We train three different multi-head classifiers: 
1) a 2-head model where each head optimizes for standard (unweighted) CE, referred to as \textbf{2HSL} (2 Heads-Single Loss); 
2) a 2-head model but with each head minimizing a differently weighed CE loss as described in section \ref{mhml}. We call this model \textbf{2HML} (2 Heads-Multi Loss));
3) Finally, we increase the number of heads to four, and we refer to this model as \textbf{4HML}.
For comparison, we include a standard single-loss one-head classifier (\textbf{SL1H}), plus models trained with Label Smoothing (\textbf{LS} \cite{muller_when_2019}), Margin-based Label Smoothing (\textbf{MbLS} \cite{liu_devil_2022}), \textbf{MixUp} \cite{thulasidasan_mixup_2019}, and using the \textbf{DCA} loss \cite{liang_improved_2020}. 
We also show the performance of Deep Ensembles (\textbf{D-Ens} \cite{lakshminarayanan_simple_2017}). 
We analyze the impact of Temperature Scaling \cite{guo_calibration_2017} in the appendix \ref{ts_app}.

\begin{afancybox}[width=.99\textwidth]
\textbf{What we expect to see:} Multi-Head Multi-Loss models should achieve a better calibration (\uline{\emph{low ECE}}) than other learning-based methods, ideally approaching Deep Ensembles calibration. 
We also expect to achieve good calibration without sacrificing predictive performance (\uline{\emph{high accuracy}}).
Both goals would be reflected jointly by a \uline{\emph{low NLL}} value, and by a \uline{\emph{better aggregated ranking}}. 
Finally we would ideally observe improved performance as we increase the diversity (comparing \textbf{2HSL} to \textbf{2HML}) and as we add heads (comparing \textbf{2HML} to \textbf{4HML}).
\end{afancybox}

\subsubsection*{Chaoyang:}
In Table \ref{chaoyang} we report the results on the Chaoyang dataset. 
Overall, accuracy is relatively low, since this dataset is challenging due to label ambiguity, and therefore calibration analysis of aleatoric uncertainty becomes meaningful here. 
As expected, we see how Deep Ensembles are the most accurate method, also with the lowest NLL, for two out of the three considered networks. 
However, we also observe noticeable differences between other learning-based calibration techniques and multi-head architectures. 
Namely, all other calibration methods achieve lower ECE than the baseline (\textbf{SL1H}) model, but \textit{at the cost of a reduced accuracy}. 
This is actually captured by NLL and rank, which become much higher for these approaches. 
In contrast, \textbf{4HML} achieves the second rank in two architectures, only behind Deep Ensembles when using a ResNet50 and a Swin-Transformer, and above any other \textbf{2HML} with a ConvNeXt, \textit{even outperforming Deep Ensembles in this case}. 
Overall, we can see a pattern: multi-loss multi-head models appear to be extremely well-calibrated (low ECE and NLL values) without sacrificing accuracy, and as we diversify the losses and increase the number of heads we tend to improve calibration.

\begin{table}[!b]
\renewcommand{\arraystretch}{1.03}
\setlength\tabcolsep{1.00pt}
\begin{center}
\begin{tabular}{c cccc cccc cccc}
& \multicolumn{4}{c}{\textbf{ResNet50}} & \multicolumn{4}{c}{\textbf{ConvNeXt}} & \multicolumn{4}{c}{\textbf{Swin-Transformer}} \\
\cmidrule(lr){2-5} \cmidrule(lr){6-9} \cmidrule(lr){10-13} &  ACC$^\uparrow$  &  ECE$_\downarrow$  &  NLL$_\downarrow$    &  Rank$_\downarrow$  &  ACC$^\uparrow$  &  ECE$_\downarrow$  &  NLL$_\downarrow$    &  Rank$_\downarrow$    &  ACC$^\uparrow$ &  ECE$_\downarrow$  &  NLL$_\downarrow$    &  Rank$_\downarrow$\\
\midrule
\textbf{SL1H}     & 80.71 & 5.79 & 53.46 & 6.0 & 81.91 & 6.94 & 50.98 & 6.3 & 83.09 & 8.73 & 52.75 & 5.0 \\
\midrule
\textbf{LS}       & 74.81 & 2.55 & 64.27 & 6.7 & 79.59 & 6.13 & 55.65 & 7.3 & 79.76 & 3.98 & 55.37 & 6.0 \\
\midrule
\textbf{MbLS}     & 75.02 & 3.26 & 63.86 & 6.7 & 79.53 & 2.94 & 53.44 & 5.3 & 80.24 & 5.06 & 54.18 & 5.7 \\
\midrule
\textbf{MixUp}    & 76.00 & 3.67 & 62.72 & 6.3 & 79.95 & 6.20 & 55.58 & 7.0 & 80.25 & 3.89 & 54.62 & 4.7 \\
\midrule
\textbf{DCA}      & 76.17 & 5.75 & 62.13 & 6.7 & 78.28 & 3.69 & 57.78 & 7.3 & 79.12 & 7.91 & 59.91 & 8.3 \\
\midrule
\midrule
\textbf{D-Ens}    & 82.19 & 2.42 & 46.64 & \ft{1.0} & 82.98 & 5.21 & 46.08 & 3.3 & 83.50 & 6.79 & 44.80 & \ft{2.7} \\
\midrule
\midrule
\textbf{2HSL}       & 80.97 & 4.36 & 51.42 & 4.0 & 81.94 & 4.30 & 46.71 & 4.3 & 82.90 & 8.20 & 54.19 & 5.7 \\
\midrule
\textbf{2HML}      & 80.28 & 4.49 & 51.86 & 5.3 & 81.97 & 3.66 & 45.96 & \sd{2.7} & 82.79 & 5.01 & 46.12 & 3.7 \\
\midrule
\textbf{4HML}      & 81.13 & 3.09 & 49.44 & \sd{2.3} & 82.17 & 1.79 & 44.73 & \ft{1.3} & 82.89 & 4.80 & 46.70 & \sd{3.3} \\
\bottomrule
\\[-0.25cm] 
\end{tabular}
\caption{Results on the \textbf{Chaoyang dataset} with different architectures and strategies.\
For each model, \unl{\textbf{best}} and \textbf{second best} ranks are marked.}\label{chaoyang}
\end{center}
\vspace{-4cm}
\end{table}

\subsubsection*{Kvasir:}
Next, we show in Table \ref{kvasir} results for the Kvasir dataset. 
Deep Ensembles again reach the highest accuracy and excellent calibration.
Interestingly, methods that smooth labels (\textbf{LS}, \textbf{MbLS}, \textbf{MixUP}) show a strong degradation in calibration and their ECE is often twice the ECE of the baseline \textbf{SL1H} model. 
We attribute this to class imbalance and the large number of categories: smoothing labels might be ineffective in this scenario.
Note that models minimizing the \textbf{DCA} loss do manage to bring the ECE down, although by giving up accuracy. 
In contrast, all multi-head models improve calibration while maintaining accuracy. 
Remarkably, \uline{\textbf{4HML} \textit{obtains lower ECE than Deep Ensembles in all cases}}. 
Also, for two out of the three architectures \textbf{4HML} ranks as the best method, and for the other one \textbf{2HML} reaches the best ranking.

\begin{table}[!t]
\renewcommand{\arraystretch}{1.03}
\setlength\tabcolsep{1.00pt}
\begin{center}
\begin{tabular}{c cccc cccc cccc}
& \multicolumn{4}{c}{\textbf{ResNet50}} & \multicolumn{4}{c}{\textbf{ConvNeXt}} & \multicolumn{4}{c}{\textbf{Swin-Transformer}} \\
\cmidrule(lr){2-5} \cmidrule(lr){6-9} \cmidrule(lr){10-13} &  ACC$^\uparrow$  &  ECE$_\downarrow$  &  NLL$_\downarrow$    &  Rank$_\downarrow$  &  ACC$^\uparrow$  &  ECE$_\downarrow$  &  NLL$_\downarrow$    &  Rank$_\downarrow$    &  ACC$^\uparrow$ &  ECE$_\downarrow$  &  NLL$_\downarrow$    &  Rank$_\downarrow$\\
\midrule
\textbf{OneH}     & 89.87 & 6.32 & 41.88 & 5.3 & 90.02 & 5.18 & 35.59 & 5.0 & 90.07 & 5.81 & 38.01 & 5.7 \\
\midrule
\textbf{LS}       & 88.13 & 14.63 & 53.96 & 7.7 & 88.24 & 6.97 & 42.09 & 6.7 & 88.74 & 9.20 & 43.46 & 8.7 \\
\midrule
\textbf{MbLS}     & 88.20 & 16.92 & 57.48 & 8.0 & 88.62 & 8.55 & 43.07 & 7.0 & 89.15 & 8.19 & 41.85 & 7.7 \\
\midrule
\textbf{MixUp}    & 87.60 & 10.28 & 50.69 & 7.3 & 87.58 & 8.96 & 48.88 & 8.7 & 89.23 & 2.11 & 35.52 & 4.3 \\
\midrule
\textbf{DCA}      & 87.14 & 3.84 & 40.50 & 6.0 & 85.27 & 4.11 & 46.78 & 7.3 & 87.62 & 4.38 & 38.44 & 7.3 \\
\midrule
\midrule
\textbf{D-Ens}    & 90.76 & 3.83 & 32.09 & 2.3 & 90.76 & 3.34 & 29.74 & 3.0 & 90.53 & 3.94 & 29.36 & 3.3 \\
\midrule
\midrule
\textbf{2HSL}       & 89.76 & 4.52 & 34.34 & 4.7 & 90.21 & 2.63 & 28.69 & \sd{2.7} & 90.40 & 3.65 & 29.14 & 3.0 \\
\midrule
\textbf{2HML}      & 90.05 & 3.62 & 31.37 & \sd{2.0} & 89.92 & 1.49 & 28.15 & \sd{2.7} & 90.19 & 2.73 & 28.66 & \sd{2.7} \\
\midrule
\textbf{4HML}      & 89.99 & 2.22 & 30.02 & \ft{1.7} & 90.10 & 1.65 & 28.01 & \ft{2.0} & 90.00 & 1.82 & 27.96 & \ft{2.3} \\
\bottomrule
\\[-0.25cm]
\end{tabular}
\caption{Results on the \textbf{Kvasir dataset} with different architectures and strategies.\
For each model, \unl{\textbf{best}} and \textbf{second best} ranks are marked.}\label{kvasir}
\end{center}
\vspace{-1cm}
\end{table}

\section{Conclusion}
Multi-Head Multi-Loss networks are classifiers with enhanced calibration and no degradation of predictive performance when compared to their single-head counterparts. 
This is achieved by simultaneously optimizing several output branches, each one minimizing a differently weighted Cross-Entropy loss. 
Weights are complementary, ensuring that each branch is rewarded for becoming specialized in a subset of the original data categories. 
Comprehensive experiments on two challenging datasets with three different neural networks show that Multi-Head Multi-Loss models consistently outperform other learning-based calibration techniques, matching and sometimes surpassing the calibration of Deep Ensembles.

\section*{Acknowledgments}
This work was supported by a Marie Sk lodowska-Curie Fellowship (No 892297) and by Australian Research Council grants (DP180103232 and FT190100525).

\bibliographystyle{splncs04}
\bibliography{miccai_PMH.bib}

\newpage
\begin{subappendices}
\renewcommand{\thesection}{\Alph{section}}%

\section{Gradient Derivations}\label{gradients}
In sections \ref{mhd} and \ref{mhml}, we referred to the gradients for the M-head model and its extension with ``deep supervision'' on each head by a weighted CE loss. 
Here we provide careful step-by-step derivation of these quantities. 

We start by going over our notation. 
We have a neural network $U_\theta$ that maps an image $\mathbf{x}$ into a vector of representations in $U_\theta(\mathbf{x})\in\mathbb{R^N}$. 
In a standard model, this feature vector would then be passed through a linear classifier, composed of a linear mapping $f:\mathbb{R^N}\rightarrow\mathbb{R^K}$ followed by a softmax operation $\sigma$. 
The intermediate vector $\z=f(U_\theta(\mathbf{x}))\in\mathbb{R^K}$ is often called logits vector, and the final vector $\p=\sigma(\z)$, whose components are $p_j=e^{-z_j}/\sum_ie^{-z_i}$, can be regarded as a ``probability vector'', indicating the likelihood of each category.

Suppose $\mathbf{x}$ has a label $y\in\{1,...,K\}$. In order to measure the classification error of the above model, we can use the Cross-Entropy (CE) loss, given by $\LL_\textrm{CE}(\p, y) = -\log(p_y)$, that is, we attempt to maximize the probability assignment at the $y$-th component in $\p$, regardless of the values elsewhere in $\p$. 
We are interested in the gradient that reaches the linear mapping in this model when we backpropagate the CE loss. 
Since this only depends on that component, we consider only the partial derivative at $p_y$, which we can find by application of the chain rule:
\begin{equation}\nonumber
\frac{\partial\LL_\textrm{CE}(p_y, y)}{\partial z_j} = \frac{\partial\LL_\textrm{CE}(p_y, y)}{\partial p_y}\cdot \frac{\partial p_y}{\partial z_j} = \frac{-1}{p_y}\cdot p_y\cdot(\delta_y^j-p_j) = p_j-\delta_y^j,
\end{equation}
where $\delta_y^j=0$ unless we are computing the partial derivative with respect to the logit of the correct class, in which case $\delta_y^j=1$. 
Therefore the complete gradient can now be written as:
\begin{equation}\label{grad_app}
\nabla_\z \LL_\textrm{CE}(\p, y) = [\,p_1,p_2,...,p_y-1, ..., p_K\,] = \p - \mathbf{y}
\end{equation}
where $\mathbf{y} = [\,0,...,1^{(k)},...0\,]\in\mathbb{R}^K$ is a one-hot representation of label $y$. 
In what follows, to avoid cumbersome notation, we will omit the point at which the partial derivatives are evaluated, hoping it will be clear from the context.

Next, we want to extend this basic architecture into a multi-head ensemble. 
In this case, the single linear classifier above is replaced by $M$ branches $f^1,..., f^M$ that map $\mathbf{x}$ into $M$ logit vectors $\z_1, ..., \z_M$, which are then passed through a softmax layer that turns them into ``probability vectors'' $\p^m=\sigma(\z^m)$. 
In this multi-head architecture, these vectors are then averaged into a single prediction $\p^\mu = \mu(\p_1,...,\p_M) = (\p_1+...+\p_M)/M$ and the loss $\LL_\textrm{CE}(\p^\mu, y)$ is computed. An example of this model is shown in Fig. \ref{fig_app1} below.

\begin{figure}[t!]
\centerline{\includegraphics[width=0.70\textwidth]{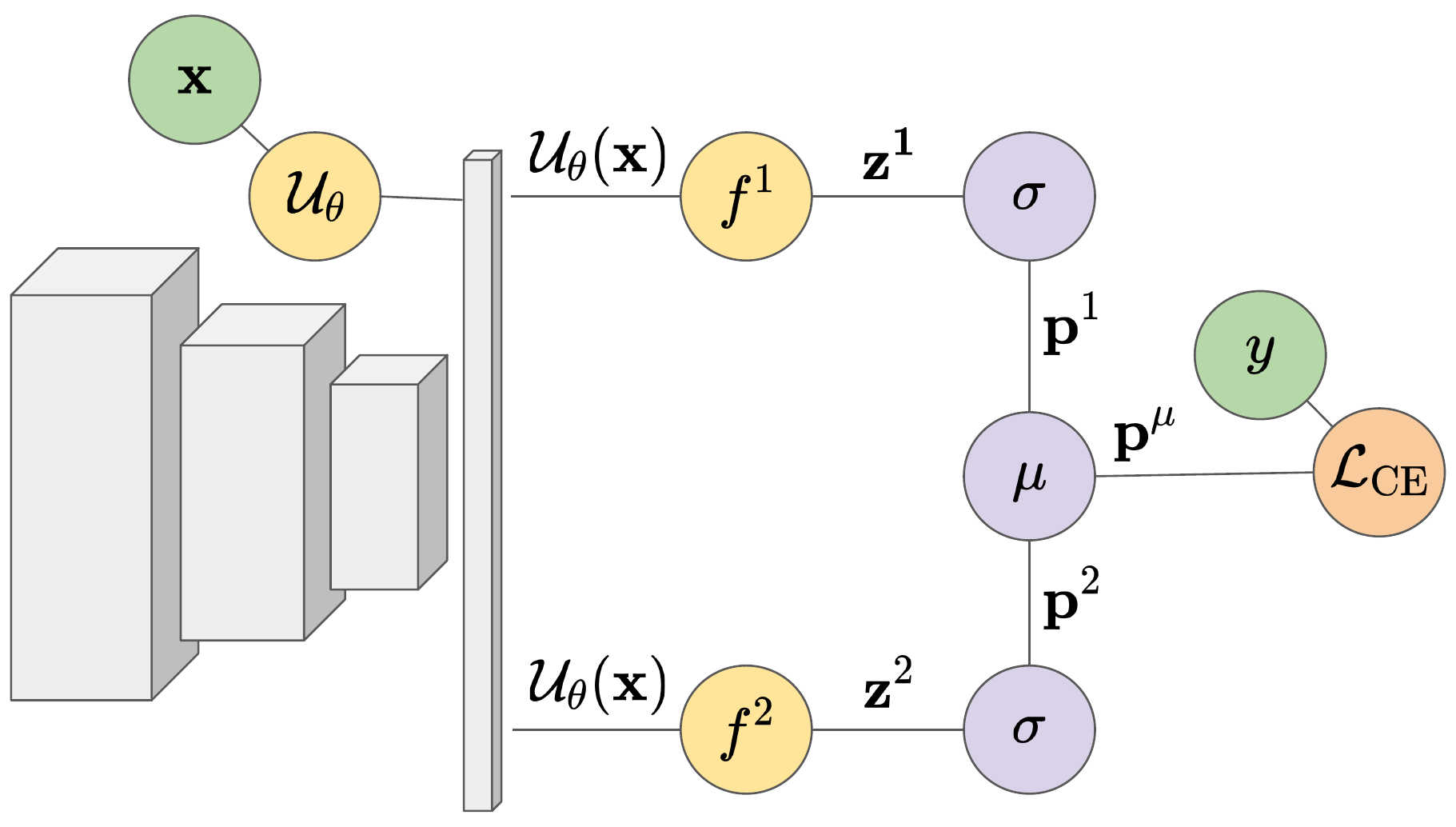}}
\caption{Schematic illustration of a multi-head model for $M$=2 heads.}
\label{fig_app1}
\end{figure}

We can now derive the gradient reaching a linear layer $f^m$ (with respect to the logits $\z^m$) when we backpropagate the loss. 
Again the loss at $\p^\mu$ only depends on the component $p_y^\mu$ of the correct category, as follows:
\begin{equation}\nonumber
\frac{\partial\LL_\textrm{CE}}{\partial z^m_j} = 
\frac{\partial\LL_\textrm{CE}}{\partial p_y^\mu}\cdot \frac{\partial p_y^\mu}{\partial p_y^m} \cdot \frac{\partial p_y^m}{\partial z_j^m}
= \frac{-1}{p_y^\mu}\cdot \frac{1}{M} \cdot p_y^m\cdot(\delta_y^j-p^m_j) = \frac{p_y^m }{\sum_i p_y^i} \cdot (p_j^m-\delta_y^j),
\end{equation}
so the gradient would be given by eq. (\ref{multi_head_grad}) in the paper, this is:
\begin{equation}\nonumber
\nabla_{\z^m} \LL_{\textrm{CE}}(\p^\mu, y) = \frac{p_y^m}{\sum_i p_y^i}(\p^m - \mathbf{y}),
\end{equation}
which tells us that the gradient in eq. (\ref{grad_app}) is scaled, for branch $f^m$, by how high is $p_y^m$ relative to the sum over all heads of the probability of the correct class.

\begin{figure}[!b]
\centerline{\includegraphics[width=0.90\textwidth]{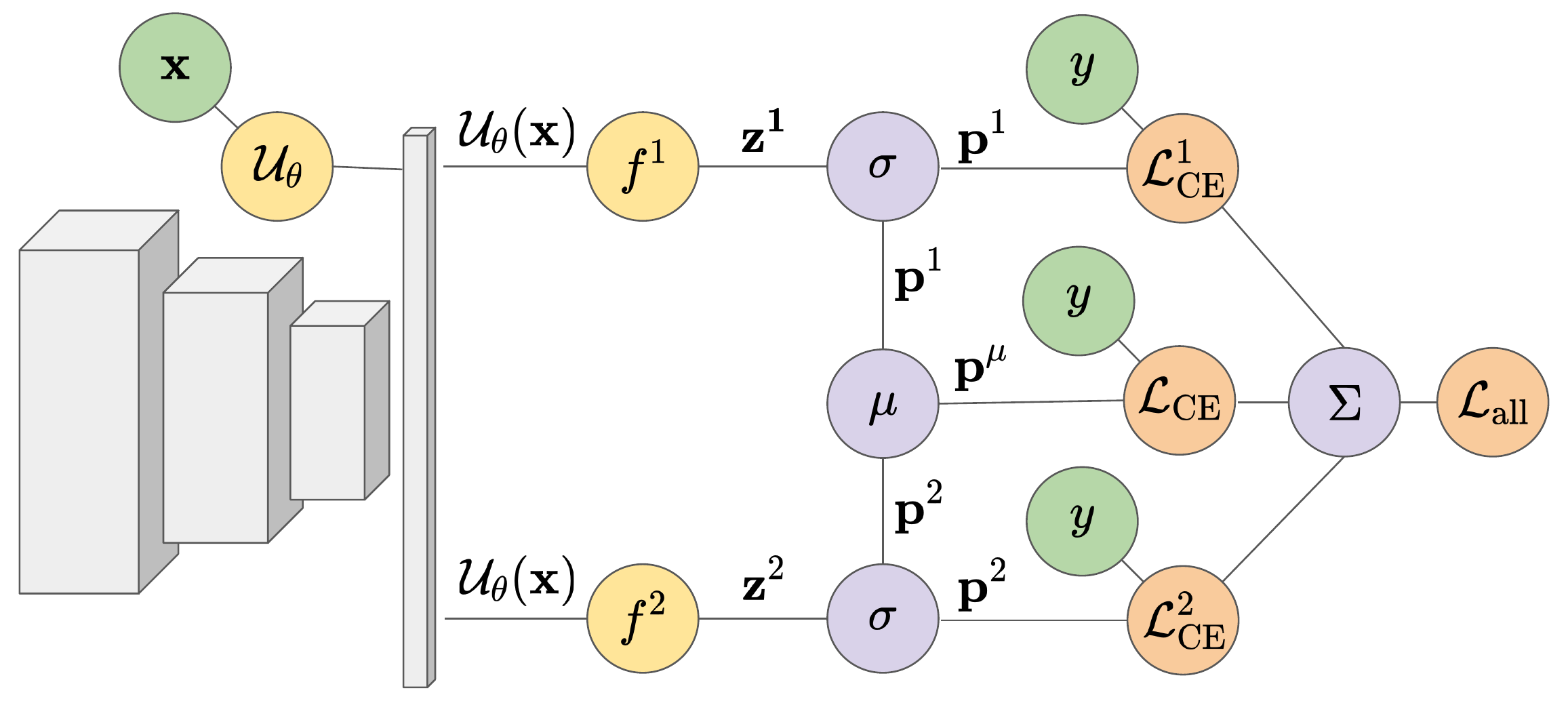}}
\caption{Multi-head model ($M$=2 heads) with additional supervision at each head.}
\label{fig_app2}
\end{figure}

In addition to using only supervision on the average prediction $\p^\mu$ by backpropagating $\LL_\textrm{CE}^\mu(p_y^\mu, y)$, we can add supervision to each individual head via an additional CE loss $\LL^m_\textrm{CE}$ at each branch's prediction $\p^m$, as shown in Fig. \ref{fig_app2}. 

In this case, the final loss to be backpropagated is the addition of all losses:
\begin{equation}\label{loss_app}
\mathcal{L}_{all}(\p, y) =  \LL_\textrm{CE}^\mu(\p^\mu, y) +\sum_m \LL^m_\textrm{CE}(\p^m, y),
\end{equation} 
where $\p=[\, \p^1,...,\p^M\,]$ is an array gathering the predictions of all $M$ heads.

We can then apply the sum rule to quickly find out the derivative of this loss at branch $f^m$ with respect to the logits $\z^m$, since only the loss on the average prediction $\p^\mu$ and the one on $\p^m$ will contribute to it:
\begin{equation}\nonumber
\frac{\partial\LL_\textrm{all}}{\partial z^m_j} = 
\frac{\partial\LL_\textrm{CE}^\mu}{\partial z^m_j} + \frac{\LL^m_\textrm{CE}}{\partial z^m_j}= (\frac{p^m_y}{\sum_i p_y^i}+1) \cdot (p_j^m-\delta_y^j),
\end{equation}
or in other words:
\begin{equation}\nonumber
\nabla_{\z^m} \mathcal{L}_{all}(\p, y) = (1+\frac{p^m_y}{\sum_i p^i_y}) (\p^m-\mathbf{y}).
\end{equation}

\begin{center}
\noindent\rule[0.5ex]{0.75\linewidth}{1pt}    
\end{center}

In the paper we use the weighted variant of the CE loss for supervising the predictions of each head.
This loss is given by $\mathcal{L}_{\boldsymbol{\omega}\textrm{-CE}}(\p,y) = - \omega_y \log(p_y)$, where we have a weight vector $\boldsymbol{\omega}\in\mathbb{R}^K$, so the greater the magnitude of a component $\omega_j$, the larger the loss assigned to mistakes in category $j$. 
The $\mathcal{L}_{\boldsymbol{\omega}\textrm{-CE}}$ loss is typically used to penalize errors in minority categories of imbalanced classification datasets. 
Since $\mathcal{L}_{\boldsymbol{\omega}\textrm{-CE}}(\p,y) = -\omega_y \LL_\textrm{CE}(\p, y)$, then 
$\displaystyle\frac{\partial}{\partial z_j}\mathcal{L}_{\boldsymbol{\omega}\textrm{-CE}}(\p,y)=\omega_y\cdot(p_j-\delta_j^y)$ and 
$\nabla_\z \mathcal{L}_{\boldsymbol{\omega}\textrm{-CE}}(\p,y)=\omega_y\cdot(\p - \mathbf{y})$.
Introducing weighted losses in eq. (\ref{loss_app}) with different vectors $\boldsymbol{\omega^m}$, our final loss function is given by:
\begin{equation}\nonumber
\LL_{\textrm{MH}}(\p, y) = \LL_{\textrm{CE}}(\p^\mu, y) + \sum_{m=1}^M \mathcal{L}_{\boldsymbol{\omega}^m\textrm{-CE}}(\p^m, y),
\end{equation}
At this point, it is easy to see that the gradient of $\LL_{\textrm{MH}}$ is indeed:
\begin{equation}\nonumber
\nabla_{\z^m} \LL_{\textrm{MH}} = \left(\omega_y^m + \frac{p_y^m}{\sum_i p_y^i}\right)(\p^\mu - \mathbf{y}).
\end{equation}

\section{Comparison to Temperature Scaling}\label{ts_app}
In this section we analyze the impact of applying temperature scaling as a post-processing step to each of the learning-based calibration methods described in the main paper. 
It should be noted that our multi-head multi-loss architecture is not directly amenable to this kind of post-processing. 
This is because Temperature Scaling operates by learning a transformation on the logits space, whereas our models are optimized over average softmax probabilities of their heads. 
This means that in our approach we first pass each heads' logits through a softmax operation and then average them. 
Attempting to find a temperature parameter for each individual head did not result in a performance improvement, and so we decided to modify the output of our model so that it would first average the logits and then pass the result through a softmax layer, which enabled temperature fitting in the logit space. 
This came at the cost of a slightly reduced performance both in terms of accuracy and calibration, since there was a reason to first apply softmax to each head and only then average: this ensures that the average is taken over similarly-scaled vectors. 
This process explains why the performance of our models in the two tables below is slightly worse than in the main paper. 
Still, it remains interesting to analyze the impact of Temperature Scaling on our models, as compared to the post-processing of other methods.

\begin{table}[!t]
\renewcommand{\arraystretch}{1.03}
\setlength\tabcolsep{1.00pt}
\begin{center}
\begin{tabular}{c cccc cccc cccc}
& \multicolumn{4}{c}{\textbf{ResNet50}} & \multicolumn{4}{c}{\textbf{ConvNeXt}} & \multicolumn{4}{c}{\textbf{Swin-Transformer}} \\
\cmidrule(lr){2-5} \cmidrule(lr){6-9} \cmidrule(lr){10-13} &  ACC$^\uparrow$  &  ECE$_\downarrow$  &  NLL$_\downarrow$    &  Rank$_\downarrow$  &  ACC$^\uparrow$  &  ECE$_\downarrow$  &  NLL$_\downarrow$  &  Rank$_\downarrow$    &  ACC$^\uparrow$ &  ECE$_\downarrow$  &  NLL$_\downarrow$    &  Rank$_\downarrow$\\
\midrule[1.15pt]
\textbf{SL1H}     & 80.71 & 5.79 & 53.46 & 4.7 & 81.91 & 6.94 & 50.98 & 5.7 & 83.09 & 8.73 & 52.75 & 4.3 \\
\midrule
\textbf{+TS}      & \textquotedbl & 2.15 & 49.96 & \sd{2.7} & \textquotedbl & 2.24 & 45.47 & 3.7 & \textquotedbl & 1.86 & 42.54 & \ft{1.3} \\
\midrule[1.15pt]
\textbf{2HSL}     & 80.24 & 5.81 & 53.40 & 5.3 & 81.92 & 5.19 & 47.31 & 4.3 & 82.74 & 6.94 & 48.73 & 4.3 \\
\midrule
\textbf{+TS}      & \textquotedbl & 2.00 & 50.82 & 3.7 & \textquotedbl & 2.41 & 45.23 & \sd{3.0} & \textquotedbl & 2.00 & 43.48 & 3.7 \\
\midrule[1.15pt]
\textbf{4HML}     & 81.15 & 4.03 & 50.71 & \sd{2.7} & 82.22 & 4.37 & 46.39 & \sd{3.0} & 82.93 & 7.86 & 51.22 & 4.3 \\
\midrule
\textbf{+TS}      & \textquotedbl & 2.67 & 49.69 & \ft{2.0} & \textquotedbl & 1.54 & 44.94 & \ft{1.3} & \textquotedbl & 2.05 & 43.42 & \sd{3.0} \\
\bottomrule
\\[-0.25cm]
\end{tabular}
\caption{Results on the \textbf{Chaoyang dataset} with/out Temperature Scaling (\textbf{+TS}).\
For each model, \unl{\textbf{best}} and \textbf{second best} ranks are marked.}\label{chaoyang_ts}
\end{center}
\vspace{-1cm}
\end{table}

\begin{table}[!b]
\renewcommand{\arraystretch}{1.03}
\setlength\tabcolsep{1.00pt}
\begin{center}
\begin{tabular}{c cccc cccc cccc}
& \multicolumn{4}{c}{\textbf{ResNet50}} & \multicolumn{4}{c}{\textbf{ConvNeXt}} & \multicolumn{4}{c}{\textbf{Swin-Transformer}} \\
\cmidrule(lr){2-5} \cmidrule(lr){6-9} \cmidrule(lr){10-13} &  ACC$^\uparrow$  &  ECE$_\downarrow$  &  NLL$_\downarrow$    &  Rank$_\downarrow$  &  ACC$^\uparrow$  &  ECE$_\downarrow$  &  NLL$_\downarrow$    &  Rank$_\downarrow$    &  ACC$^\uparrow$ &  ECE$_\downarrow$  &  NLL$_\downarrow$    &  Rank$_\downarrow$\\
\midrule
\textbf{SL1H}     & 89.87 & 6.32 & 41.88 & 5.7 & 90.02 & 5.18 & 35.59 & 5.0 & 90.02 & 5.18 & 35.59 & 5.0 \\
\midrule
\textbf{+TS}      & \textquotedbl & 1.80 & 32.01 & 4.3 & \textquotedbl & 1.48 & 29.26 & 4.0 & \textquotedbl & 1.48 & 29.26 & 4.0 \\
\midrule[1.15pt]
\textbf{2HSL}     & 90.07 & 4.16 & 32.12 & 3.7 & 89.90 & 1.94 & 28.37 & 4.7 & 89.90 & 1.94 & 28.37 & 4.7 \\
\midrule
\textbf{+TS}      & \textquotedbl & 1.45 & 29.82 & \ft{1.3} & \textquotedbl & 1.45 & 28.17 & 3.3 & \textquotedbl & 1.45 & 28.17 & 3.3 \\
\midrule[1.15pt]
\textbf{4HML}     & 90.00 & 3.48 & 31.21 & 3.3 & 90.11 & 1.87 & 28.20 & \sd{2.7} & 90.11 & 1.87 & 28.20 & \sd{2.7} \\
\midrule
\textbf{+TS}      & \textquotedbl & 1.71 & 30.11 & \sd{2.7} & \textquotedbl & 1.30 & 27.97 & \ft{1.3} & \textquotedbl & 1.30 & 27.97 & \ft{1.3} \\
\bottomrule
\\[-0.25cm]
\end{tabular}
\caption{Results on the \textbf{Kvasir dataset} with/out Temperature Scaling.\
For each model, \unl{\textbf{best}} and \textbf{second best} ranks are marked.}\label{kvasir_ts}
\end{center}
\vspace{-1cm}
\end{table}

Tables \ref{chaoyang_ts} and \ref{kvasir_ts} show results for the Chaoyang and Kvasir datasets when considering a single-head model and multi-head counterparts trained with the proposed multi-loss strategy. 
We add the result of calibrating the temperature of each model just below the unprocessed probabilities for an easy comparison of post-processing impact. 
We can quikcly see that regardless of the disadvantage, explained above, that Multi-Loss Multi-Head models face when adding a post-hoc calibration layer, they are still a better choice over a standard one-head model with tempered probabilities. 
In the Chaoyang dataset, the four-head model \textbf{4HML} achieves the best average ranking for two of the three backbone architectures, and the second average ranking for the other one, and the same happens with the Kvasir dataset. 
Noticeably, the \textbf{4HML} architecture was already well-calibrated prior to any post-processing. 
In both datasets, the non-tempered \textbf{4HML} probabilities had the second average rank in two occasions, only improved by its own temperature calibration.

\section{Further Experimental Results}\label{more_results}
In the main paper we reported results without dispersion measures to save space. Here we provide expanded tables that contain standard deviation over 5 runs of experiments.
In addition, we add results for PathMNIST \cite{yang_medmnist_2021}, a simple dataset containing 107,180 $28\times28$ histopathological colon images evenly distributed in nine classes.
Because this is a relatively easy dataset, most methods achieve a similar, high accuracy, which obfuscates a bit the rankings in Table \ref{pathmnist_dispersion}. 
Nonetheless, we can still appreciate how the observations made in section \ref{perf_sec} hold also here: both of our models (\textbf{2HML} and \textbf{4HML}) are among the top performers in terms of ECE and NLL, rivaling Deep Ensembles, with \textbf{4HML} scoring most of the times above its two-headed counterpart.

\begin{sidewaystable}
\renewcommand{\arraystretch}{1.03}
\setlength\tabcolsep{1.00pt}
\centering
\caption{Results on the \textbf{Chaoyang dataset}, with standard deviation for 5 training runs.}\label{chaoyang_dispersion}
\smallskip
\begin{tabular}{c cccc cccc cccc}
& \multicolumn{4}{c}{\textbf{ResNet50}} & \multicolumn{4}{c}{\textbf{ConvNeXt}} & \multicolumn{4}{c}{\textbf{Swin-Transformer}} \\
\cmidrule(lr){2-5} \cmidrule(lr){6-9} \cmidrule(lr){10-13} &  ACC$^\uparrow$  &  ECE$_\downarrow$  &  NLL$_\downarrow$    &  Rank$_\downarrow$  &  ACC$^\uparrow$  &  ECE$_\downarrow$  &  NLL$_\downarrow$    &  Rank$_\downarrow$    &  ACC$^\uparrow$ &  ECE$_\downarrow$  &  NLL$_\downarrow$    &  Rank$_\downarrow$\\
\midrule
\textbf{SL1H}     & 80.71$\pm$0.10 & 5.79$\pm$0.64 & 53.46$\pm$1.75 & 6.0 & 81.91$\pm$0.24 & 6.94$\pm$0.46 & 50.98$\pm$1.49 & 6.3 & 83.09$\pm$0.13 & 8.73$\pm$0.51 & 52.75$\pm$0.45 & 5.0 \\
\midrule
\textbf{LS}       & 74.81$\pm$0.40 & 2.55$\pm$1.88 & 64.27$\pm$0.49 & 6.7 & 79.59$\pm$0.34 & 6.13$\pm$0.62 & 55.65$\pm$2.17 & 7.3 & 79.76$\pm$0.41 & 3.98$\pm$0.73 & 55.37$\pm$0.88 & 6.0 \\
\midrule
\textbf{MbLS}     & 75.02$\pm$0.54 & 3.26$\pm$1.97 & 63.86$\pm$0.97 & 6.7 & 79.53$\pm$0.26 & 2.94$\pm$0.93 & 53.44$\pm$0.76 & 5.3 & 80.24$\pm$0.19 & 5.06$\pm$0.73 & 54.18$\pm$1.65 & 5.7 \\
\midrule
\textbf{MixUp}    & 76.00$\pm$0.42 & 3.67$\pm$1.51 & 62.72$\pm$0.73 & 6.3 & 79.95$\pm$0.33 & 6.20$\pm$0.94 & 55.58$\pm$2.28 & 7.0 & 80.25$\pm$0.33 & 3.89$\pm$0.54 & 54.62$\pm$0.71 & 4.7 \\
\midrule
\textbf{DCA}      & 76.17$\pm$0.33 & 5.75$\pm$0.87 & 62.13$\pm$1.68 & 6.7 & 78.28$\pm$0.19 & 3.69$\pm$0.62 & 57.78$\pm$0.73 & 7.3 & 79.12$\pm$0.28 & 7.91$\pm$0.79 & 59.91$\pm$1.69 & 8.3 \\
\midrule
\midrule
\textbf{D-Ens}    & 82.19 & 2.42 & 46.64 & 1.0 & 82.98 & 5.21 & 46.08 & 3.3 & 83.50 & 6.79 & 44.80 & 2.7 \\
\midrule
\midrule
\textbf{2HSL}       & 80.97$\pm$0.28 & 4.36$\pm$0.87 & 51.42$\pm$1.88 & 4.0 & 81.94$\pm$0.21 & 4.30$\pm$0.29 & 46.71$\pm$1.17 & 4.3 & 82.90$\pm$0.21 & 8.20$\pm$0.63 & 54.19$\pm$1.76 & 5.7 \\
\midrule
\textbf{2HML}      & 80.28$\pm$0.26 & 4.49$\pm$0.81 & 51.86$\pm$1.29 & 5.3 & 81.97$\pm$0.19 & 3.66$\pm$0.31 & 45.96$\pm$1.15 & 2.7 & 82.79$\pm$0.22 & 5.01$\pm$0.43 & 46.12$\pm$1.24 & 3.7 \\
\midrule
\textbf{4HML}      & 81.13$\pm$0.17 & 3.09$\pm$0.82 & 49.44$\pm$0.76 & 2.3 & 82.17$\pm$0.10 & 1.79$\pm$0.21 & 44.73$\pm$0.30 & 1.3 & 82.89$\pm$0.26 & 4.80$\pm$0.44 & 46.70$\pm$1.58 & 3.3 \\
\bottomrule
\\[-0.25cm]
\end{tabular}

\bigskip\bigskip  %
\caption{Results on the \textbf{Kvasir dataset} , with standard deviation for 5 training runs.}\label{kvasir_dispersion}
\smallskip
\smallskip
\begin{tabular}{c cccc cccc cccc}
& \multicolumn{4}{c}{\textbf{ResNet50}} & \multicolumn{4}{c}{\textbf{ConvNeXt}} & \multicolumn{4}{c}{\textbf{Swin-Transformer}} \\
\cmidrule(lr){2-5} \cmidrule(lr){6-9} \cmidrule(lr){10-13} &  ACC$^\uparrow$  &  ECE$_\downarrow$  &  NLL$_\downarrow$    &  Rank$_\downarrow$  &  ACC$^\uparrow$  &  ECE$_\downarrow$  &  NLL$_\downarrow$    &  Rank$_\downarrow$    &  ACC$^\uparrow$ &  ECE$_\downarrow$  &  NLL$_\downarrow$    &  Rank$_\downarrow$\\
\midrule
\textbf{SL1H}     & 89.87$\pm$0.16 & 6.32$\pm$0.18 & 41.88$\pm$0.24 & 5.3 & 90.02$\pm$0.12 & 5.18$\pm$0.27 & 35.59$\pm$0.52 & 5.0 & 90.07$\pm$0.05 & 5.81$\pm$0.43 & 38.01$\pm$0.72 & 5.7 \\
\midrule
\textbf{LS}       & 88.13$\pm$1.05 & 14.63$\pm$0.46 & 53.96$\pm$1.99 & 7.7 & 88.24$\pm$0.80 & 6.97$\pm$0.56 & 42.09$\pm$1.60 & 6.7 & 88.74$\pm$1.18 & 9.20$\pm$0.68 & 43.46$\pm$1.77 & 8.7 \\
\midrule
\textbf{MbLS}     & 88.20$\pm$1.61 & 16.92$\pm$0.65 & 57.48$\pm$1.05 & 8.0 & 88.62$\pm$0.78 & 8.55$\pm$0.22 & 43.07$\pm$2.06 & 7.0 & 89.15$\pm$0.63 & 8.19$\pm$0.49 & 41.85$\pm$0.45 & 7.7 \\
\midrule
\textbf{MixUp}    & 87.60$\pm$0.55 & 10.28$\pm$0.50 & 50.69$\pm$2.41 & 7.3 & 87.58$\pm$0.24 & 8.96$\pm$0.58 & 48.88$\pm$2.81 & 8.7 & 89.23$\pm$0.26 & 2.11$\pm$0.32 & 35.52$\pm$0.25 & 4.3 \\
\midrule
\textbf{DCA}      & 87.14$\pm$0.41 & 3.84$\pm$0.64 & 40.50$\pm$0.76 & 6.0 & 85.27$\pm$0.45 & 4.11$\pm$0.89 & 46.78$\pm$0.94 & 7.3 & 87.62$\pm$0.31 & 4.38$\pm$0.82 & 38.44$\pm$1.62 & 7.3 \\
\midrule
\midrule
\textbf{D-Ens}    & 90.76 & 3.83 & 32.09 & 2.3 & 90.76 & 3.34 & 29.74 & 3.0 & 90.53 & 3.94 & 29.36 & 3.3 \\
\midrule
\midrule
\textbf{2HSL}       & 89.76$\pm$0.14 & 4.52$\pm$0.27 & 34.34$\pm$0.93 & 4.7 & 90.21$\pm$0.09 & 2.63$\pm$0.14 & 28.69$\pm$0.45 & 2.7 & 90.40$\pm$0.04 & 3.65$\pm$0.17 & 29.14$\pm$0.67 & 3.0 \\
\midrule
\textbf{2HML}      & 90.05$\pm$0.16 & 3.62$\pm$0.40 & 31.37$\pm$0.78 & 2.0 & 89.92$\pm$0.10 & 1.49$\pm$0.31 & 28.15$\pm$0.28 & 2.7 & 90.19$\pm$0.04 & 2.73$\pm$0.33 & 28.66$\pm$0.64 & 2.7 \\
\midrule
\textbf{4HML}      & 89.99$\pm$0.15 & 2.22$\pm$0.25 & 30.02$\pm$0.53 & 1.7 & 90.10$\pm$0.22 & 1.65$\pm$0.29 & 28.01$\pm$0.42 & 2.0 & 90.00$\pm$0.07 & 1.82$\pm$0.32 & 27.96$\pm$0.35 & 2.3 \\
\bottomrule
\\[-0.25cm]
\end{tabular}
\end{sidewaystable}

\newpage

\begin{sidewaystable}
\renewcommand{\arraystretch}{1.03}
\setlength\tabcolsep{1.00pt}
\centering
\caption{Results on the \textbf{PathMnist dataset}, with standard deviation for 5 training runs.}\label{pathmnist_dispersion}
\smallskip
\begin{tabular}{c cccc cccc cccc}
& \multicolumn{4}{c}{\textbf{ResNet50}} & \multicolumn{4}{c}{\textbf{ConvNeXt}} & \multicolumn{4}{c}{\textbf{Swin-Transformer}} \\
\cmidrule(lr){2-5} \cmidrule(lr){6-9} \cmidrule(lr){10-13} &  ACC$^\uparrow$  &  ECE$_\downarrow$  &  NLL$_\downarrow$    &  Rank$_\downarrow$  &  ACC$^\uparrow$  &  ECE$_\downarrow$  &  NLL$_\downarrow$    &  Rank$_\downarrow$    &  ACC$^\uparrow$ &  ECE$_\downarrow$  &  NLL$_\downarrow$    &  Rank$_\downarrow$\\
\midrule
\textbf{SL1H}     & 89.88$\pm$0.05 & 5.42$\pm$0.26 & 33.36$\pm$0.21 & 8.7 & 93.36$\pm$0.03 & 3.34$\pm$0.26 & 22.91$\pm$0.22 & 6.0 & 92.10$\pm$0.03 & 5.00$\pm$0.09 & 30.91$\pm$0.09 & 7.7 \\
\midrule
\textbf{LS}       & 90.15$\pm$0.13 & 3.72$\pm$0.14 & 32.50$\pm$0.21 & 6.7 & 93.45$\pm$0.07 & 3.10$\pm$0.27 & 23.69$\pm$0.16 & 6.0 & 92.58$\pm$0.06 & 2.89$\pm$0.22 & 26.39$\pm$0.26 & 2.7 \\
\midrule
\textbf{MbLS}     & 90.16$\pm$0.03 & 2.61$\pm$0.22 & 31.24$\pm$0.21 & 4.7 & 93.47$\pm$0.02 & 1.81$\pm$0.14 & 22.87$\pm$0.21 & 3.3 & 92.56$\pm$0.06 & 2.26$\pm$0.23 & 25.96$\pm$0.25 & 2.0 \\
\midrule
\textbf{MixUp}    & 89.81$\pm$0.07 & 2.49$\pm$0.21 & 31.81$\pm$0.54 & 6.0 & 92.81$\pm$0.04 & 1.45$\pm$0.21 & 24.14$\pm$0.17 & 6.0 & 91.98$\pm$0.05 & 1.21$\pm$0.21 & 26.23$\pm$0.13 & 4.3 \\
\midrule
\textbf{DCA}      & 90.14$\pm$0.05 & 4.54$\pm$0.27 & 31.10$\pm$0.28 & 6.7 & 92.76$\pm$0.02 & 2.56$\pm$0.17 & 21.56$\pm$0.09 & 5.7 & 92.20$\pm$0.03 & 4.33$\pm$0.20 & 27.26$\pm$0.21 & 5.3 \\
\midrule
\midrule
\textbf{D-Ens}    & 90.77 & 3.46 & 27.52 & 2.7 & 93.76 & 2.44 & 20.18 & 2.0 & 92.23 & 4.25 & 26.40 & 4.3 \\
\midrule
\midrule
\textbf{2HSL}       & 90.55$\pm$0.04 & 3.78$\pm$0.33 & 28.84$\pm$0.12 & 4.3 & 93.07$\pm$0.03 & 3.44$\pm$0.19 & 23.45$\pm$0.20 & 7.7 & 92.05$\pm$0.04 & 4.92$\pm$0.26 & 30.62$\pm$0.18 & 7.3 \\
\midrule
\textbf{2HML}      & 90.54$\pm$0.04 & 2.82$\pm$0.33 & 28.32$\pm$0.32 & 3.3 & 93.17$\pm$0.02 & 2.89$\pm$0.17 & 22.39$\pm$0.20 & 5.0 & 92.04$\pm$0.04 & 4.34$\pm$0.19 & 28.77$\pm$0.27 & 7.0 \\
\midrule
\textbf{4HML}      & 90.47$\pm$0.06 & 1.23$\pm$0.14 & 27.46$\pm$0.22 & 2.0 & 93.14$\pm$0.03 & 1.74$\pm$0.24 & 21.17$\pm$0.28 & 3.3 & 92.01$\pm$0.03 & 2.96$\pm$0.23 & 25.80$\pm$0.40 & 4.3 \\
\bottomrule
\\[-0.25cm]
\end{tabular}
\end{sidewaystable}
\end{subappendices}

\end{document}